\documentclass[runningheads]{llncs}
\usepackage{biblatex}  
\usepackage[T1]{fontenc}
\usepackage{graphicx}
\usepackage{color}
\usepackage{bbm}
\usepackage{amsmath}
\usepackage{multirow}
\usepackage{multicol}
\usepackage{amssymb}
\usepackage{pifont}

\begin{document}
\title{MRN: Harnessing 2D Vision Foundation Models for Diagnosing Parkinson's Disease with Limited 3D MR Data}
\titlerunning{MRN}
%
\author{
Shaodong Ding\inst{1}\orcidID{0009-0006-9768-4219} \and
Ziyang Liu\inst{2}\orcidID{0000-0001-6828-7587} \and
Yijun Zhou\inst{1} \and
Tao Liu$^*$\inst{1}\orcidID{0000-0002-7783-3073}
}
\authorrunning{Ding et al.}
%
\institute{Beijing Advanced Innovation Center for Biomedical Engineering, School of Biological Science and Medical Engineering, Beihang University, Beijing, China. \and
Beijing Celebrain Technology Co., Ltd., Beijing 100083, China.
\email{$^*$tao\_liu@buaa.edu.cn}
}
\maketitle              
\begin{abstract}

The automatic diagnosis of Parkinson’s disease is in high clinical demand due to its prevalence and the importance of targeted treatment.
Current clinical practice often relies on diagnostic biomarkers in QSM and NM-MRI images. 
However, the lack of large, high-quality datasets makes training diagnostic models from scratch prone to overfitting.
Adapting pre-trained 3D medical models is also challenging, 
as the diversity of medical imaging leads to mismatches in voxel spacing and modality between pre-training and fine-tuning data.
In this paper, we address these challenges by leveraging 2D vision foundation models (VFMs). 
Specifically, we crop multiple key ROIs from NM and QSM images, 
process each ROI through separate branches to compress the ROI into a token, 
and then combine these tokens into a unified patient representation for classification. 
Within each branch, we use 2D VFMs to encode axial slices of the 3D ROI volume and fuse them into the ROI token, 
guided by an auxiliary segmentation head that steers the feature extraction toward specific brain nuclei.
Additionally, we introduce multi-ROI supervised contrastive learning, 
which improves diagnostic performance by pulling together representations of patients from the same class while pushing away those from different classes.
Our approach achieved first place in the MICCAI 2025 PDCADxFoundation challenge, 
with an accuracy of 86.0\% trained on a dataset of only 300 labeled QSM and NM-MRI scans, 
outperforming the second-place method by 5.5\%.
These results highlight the potential of 2D VFMs for clinical analysis of 3D MR images.


\end{abstract}
\section{Introduction}




Parkinson's disease (PD) is a neurodegenerative movement disorder that progressively affects mood, mobility, sleep, and both motor and non-motor functions, 
impacting more than ten million people worldwide~\cite{rabie2025review}. 
Early diagnosis is critical, as it improves prognosis and long-term disease management. 
Although typical symptoms such as shaking or tremor, unsteady balance, and slowing of body movements,
can aid in diagnosis, 
early-stage PD is often difficult to distinguish from other disorders with overlapping motor and non-motor symptoms~\cite{wang2023automatic}. 
Therefore, accurately differentiating early-stage PD is essential for effective treatment and improved patient outcomes.

Recent advances in multi-parametric magnetic resonance imaging (MRI), including quantitative susceptibility mapping (QSM) and neuromelanin-sensitive (NM) imaging, 
have enabled the extraction of complementary biomarkers from deep brain nuclei for early PD diagnosis~\cite{he2021imaging}. 
However, effective tools for automated PD diagnosis (APD) remain limited. 
Conventional approaches rely on manual delineation of brain nuclei, 
such as the substantia nigra (SN) and its pars compacta (SNpc),
which is repetitive and time-consuming. 
Moreover, the complex structural and intensity patterns within these nuclei are difficult to fully exploit by radiologists, 
making APD a significant clinical challenge.

Currently, two main approaches are used for APD with MRI: radiomics-based methods and deep learning–based methods.
Radiomics approaches extract quantitative features from manually delineated brain nuclei and use machine learning models such as linear regression (LR), support vector machines (SVM), or extreme gradient boosting (XGBoost) for classification~\cite{xiao2019quantitative, kang2022combining}. 
Although these methods can achieve high accuracy, 
they are typically trained on small datasets (fewer than 200 samples) and usually rely only on QSM images.
Deep learning–based methods, in contrast, automatically learn discriminative PD patterns from MRI scans~\cite{wang2023automatic}. 
For example, Wang et al. developed a convolutional neural network (CNN) using 376 paired QSM and T1w images, achieving an area under the receiver opearating curve (AUC) of 0.845 on an external cohort of 83 PD and 72 healthy control (HC) subjects. 
However, neither radiomics nor deep learning methods have yet fully exploited the complementary information from both QSM and NM images.
Among the two, deep learning methods hold greater promise than radiomics~\cite{xiao2019quantitative}, 
as they can scale to millions of images and potentially learn features that generalize better to unseen data. 
Nevertheless, progress has been limited due to the lack of large public datasets for APD.

Adapting vision foundation models (VFMs) pretrained on massive datasets~\cite{oquab2023dinov2, simeoni2025dinov3, bai2024m3d, wu2024voco} to the APD task is promising, 
given the limited availability of medical data due to acquisition difficulty and privacy policies.
Intuitively, 3D VFMs pretrained on MR images might seem most suitable for APD, 
which also relies on QSM and NM-MRI. 
However, differences in scanner parameters and imaging modalities introduce substantial variability in intensity histograms, voxel spacing, and tissue contrast~\cite{yoon2024domain}. 
This creates significant misalignment between pretraining data (e.g., T1w, T2, FLAIR with isotropic 1 mm voxel spacing) and fine-tuning data (QSM, NM with anisotropic voxel spacing). 
Because deep learning models are highly sensitive to such input variations, directly fine-tuning 3D VFMs becomes non-trivial (see Sec.~\ref{sec:discussion}).
In contrast, 2D VFMs pretrained on natural images offer key advantages. 
Natural images have a fixed grayscale range (256 levels), 
making it straightforward to map MR intensities to this scale and thus reduce modality-related intensity misalignment. 
Moreover, 2D VFMs operate on slices rather than requiring isotropic volumes, 
naturally avoiding the voxel spacing inconsistencies.
Therefore, it is feasible to adapting a 2D VFMs to APD task in terms of the above reasons.


In this paper, we propose a novel method for APD by adapting 2D VFMs.
Specifically, we introduce \underline{M}ulti-\underline{R}OI driven classification \underline{N}etwork (MRN), 
which processes four ROIs cropped from paired NM and QSM images in parallel. 
Each ROI is compressed into a token, 
and these tokens are combined into a unified patient-level representation for classification. 
The selected ROIs cover key brain nuclei, 
encouraging MRN to base its predictions on clinically relevant regions.
Within each ROI branch, we further propose \underline{R}OI feature \underline{E}xtraction and \underline{s}egmentation (RES). 
RES leverages 2D VFMs to extract features from axial slices of each 3D ROI volume and fuses them into the ROI token. 
An auxiliary segmentation head is incorporated to guide feature extraction,
ensuring RES focuses on the brain nuclei within each ROI.
Finally, we employ multi-ROI supervised contrastive learning, 
which pulls together patient-level representations from the same class and pushes away those from different classes. 
This enhances the consistency between cropped ROIs and improves classification accuracy.
We evaluate MRN on the MICCAI 2025 PDCADxFoundation challenge, 
where it achieved first place with an accuracy of 86.0\%, 
outperforming competing methods by a substantial margin.
Our contributions can be summarized as follows:
\begin{itemize}
    \item We propose \underline{M}ulti-\underline{R}OI driven classification \underline{N}etwork (MRN) for PD/HC classification using four ROIs cropped from paired NM and QSM images\footnote{Code will be available at https://github.com/dongdongtong/MICCAI25-PDCADxFoundation-MRN}. 
    MRN processes the ROIs in parallel branches, 
    compresses each into a token, 
    and integrates them into a unified patient representation for classification.
    \item We introduce \underline{R}OI feature \underline{E}xtraction and \underline{s}egmentation (RES) as the core component of each MRN branch. RES compresses the global information of a 3D ROI volume into a token using 2D pretrained VFMs, 
    while an auxiliary segmentation head guides feature extraction to focus on brain nuclei.
    \item We propose multi-ROI Supervised Momentum Contrastive learning (mSupMoCo) to further enhance classification. 
    mSupMoCo learns clustered patient representations by pulling together patient-level representations of the same class and pushing apart those of different classes.
    \item We validate MRN with RES in the MICCAI 2025 PDCADxFoundation challenge, 
    achieving first place with an accuracy of 86.0\%, surpassing the second-place method by about 6.0\%. 
    These results demonstrate the strong potential of adapting 2D VFMs pretrained on natural images for the analysis of 3D volumetric medical data.
\end{itemize}

\section{Related Work}

\subsection{Automated Parkinson's Disease Diagnosis}
There are hundreds of studies that have investigated APD using various modalities such as imaging biomarkers, motor symptom monitoring, and voice analysis~\cite{rabie2025review}. 
In this work, we focus specifically on machine learning and deep learning–based approaches that utilize MR images.

Existing methods can be broadly divided into two categories based on the type of MR data they employ.
The first group leverages T2 images from the public PPMI dataset~\cite{marek2018parkinson}. 
To reduce the data requirements of CNNs, these studies apply transfer learning with pretrained networks such as AlexNet on whole-brain images,
either directly~\cite{sivaranjini2020deep} or with GAN-augmented data~\cite{kaur2021diagnosis}.
However, these methods are limited to 2D architectures, 
which cannot capture depth information, 
and they lack explicit priors on SN ROIs.
The second group adopts QSM images and focuses on specific brain nuclei for APD~\cite{li20193d, xiao2019quantitative, zhang2022histogram, kang2022combining, wang2023automatic}.
Within this group, radiomics-based approaches extract features from SN nuclei and train traditional models such as SVMs or XGBoost, 
reporting strong diagnostic performance~\cite{li20193d, kang2022combining, zhang2022histogram}.
Deep learning–based approaches crop ROIs from QSM images and employ CNNs for classification~\cite{xiao2019quantitative, wang2023automatic}.
Among all studies, only~\cite{xiao2019quantitative, wang2023automatic} utilize 3D CNNs to process full ROI volumes.
Nevertheless, CNNs are highly data-dependent, and the limited dataset sizes used in these works increase the risk of overfitting.
Unlike previous methods, our approach processes multiple 3D ROIs from both QSM and NM images simultaneously by leveraging 2D VFMs. 
This design reduces the data requirements while achieving superior diagnostic performance.








\subsection{Vision Foundation Models}
VFMs are large-scale vision models trained with millions or even billions of images, 
either 2D natural images~\cite{radford2021learning, oquab2023dinov2, fang2023eva, simeoni2025dinov3, ravi2024sam} or 3D medical images~\cite{bai2024m3d, he2025vista3d, wu2024voco, du2024segvol}, 
using self-supervised~\cite{oquab2023dinov2, simeoni2025dinov3, wu2024voco}, 
weakly supervised~\cite{radford2021learning, fang2023eva, bai2024m3d}, 
or fully supervised learning~\cite{ravi2024sam, du2024segvol}.
Generally, VFMs are not limited to solve only single task like segmentation~\cite{ravi2024sam, he2025vista3d, du2024segvol} or classification~\cite{radford2021learning, bai2024m3d},
but can be adapted to a range of tasks after finetuning~\cite{wu2024voco}.
For example, VoCo~\cite{wu2024voco} and its extended version~\cite{wu2024large} have demonstrated versatility across tasks such as brain tumor segmentation, 
lung nodule classification, 
and vision–language retrieval. 
However, most downstream tasks in VoCo rely on data modalities already included in the pretraining set, 
leaving performance on unseen modalities and tasks largely unexplored.
Similarly, M3D-CLIP~\cite{bai2024m3d}, pretrained mainly on CT images, faces uncertainty when adapted to unseen modalities, such as MRI.
Moreover, existing medical VFMs generally assume isotropic voxel spacing, 
while real-world clinical MR and CT images often have heterogeneous spacings and sequences, limiting their applicability in small-scale clinical tasks. 
In contrast, our MRN directly leverages 2D VFMs, which operate on slices rather than volumes, making them naturally agnostic to voxel spacing and image modalities.

\subsection{Contrastive Learning}
Contrastive learning can be broadly categorized into two paradigms: unsupervised and supervised.
Unsupervised contrastive learning is essentially instance discrimination, 
where the goal is to learn representations such that semantically similar images are implicitly clustered in the latent space~\cite{he2020momentum, chen2021exploring}.
In contrast, supervised contrastive learning explicitly organizes images into class-specific clusters, 
pulling together samples from the same class while pushing apart those from different classes~\cite{khosla2020supervised, wang2021exploring, ding2025c3r}.
In this work, we take a different perspective: instead of treating a single image as the contrastive unit, 
we concatenate multiple ROIs from a patient’s paired NM and QSM images to form a unified patient-level representation,
which serves as the contrastive sample for supervised contrastive learning.

\begin{figure}
    \centering
    \includegraphics[width=1.0\linewidth]{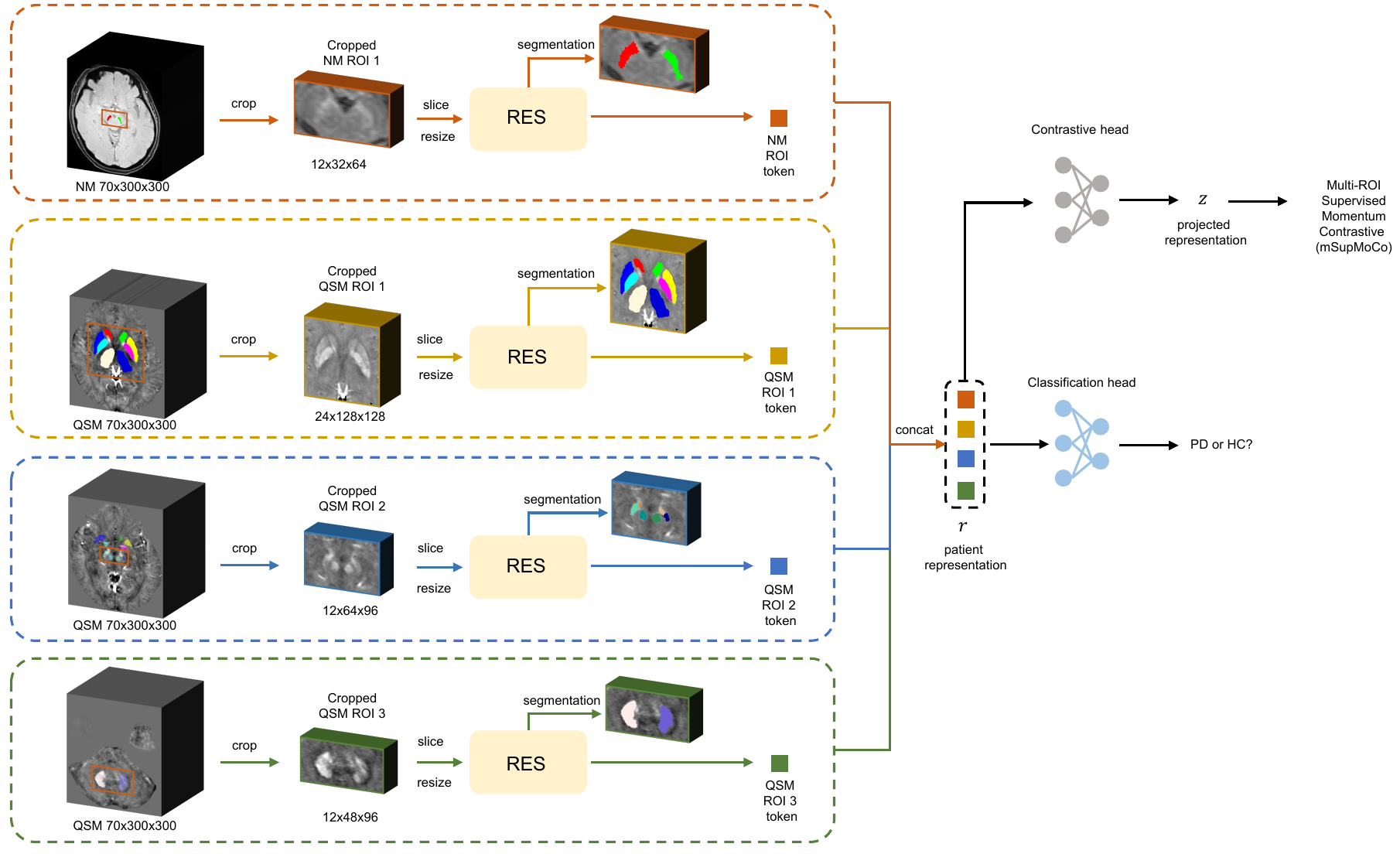}
    \caption{
    Overview of the \underline{M}ulti-\underline{R}OI driven classification \underline{N}etwork (MRN) for classifying Parkinson’s disease (PD) versus healthy controls (HC). The framework consists of a main classification task and two auxiliary tasks: 1) ROI segmentation and 2) multi-ROI supervised momentum contrastive learning (mSupMoCo).
    }
    \label{fig:mrn}
\end{figure}

\section{Method}

Firstly, we give an overview of our classification framework in Sec.~\ref{sec:overview}.
Next, we detail the feature extraction process of the framework named RES, in Sec.~\ref{sec:res},
and classification Network MRN in Sec.~\ref{sec:mrn}. 
Lastly, we summarize the training objectives in Sec.~\ref{sec:losses}.

\subsection{Overview}
\label{sec:overview}

Our framework takes a patient’s paired NM and QSM images, along with the corresponding brain nuclei segmentation mask, as input to classify the patient as HC or PD.
As illustrated in Fig.~\ref{fig:mrn}, MRN is designed as a multi-task learning framework consisting of one main classification task and two auxiliary tasks: ROI segmentation and supervised contrastive learning.
MRN contains four parallel RES branches (Fig.~\ref{fig:res}), 
each responsible for extracting a global ROI token and segmenting brain nuclei from one NM ROI and three QSM ROIs. 
The auxiliary segmentation task guides RES to focus on disease-relevant information within specific regions. 
The four ROI tokens are then concatenated into a unified patient-level representation.
Furthermore, we apply mSupMoCo~\cite{ding2025c3r, he2020momentum}, 
which pulls together patient-level representations from the same class and pushes apart those from different classes, 
thereby enhancing feature discriminability. 
Finally, the refined patient representation is used for classification.

\subsection{ROI Feature Extraction and Segmentation}
\label{sec:res}

\begin{figure}
    \centering
    \includegraphics[width=1.0\linewidth]{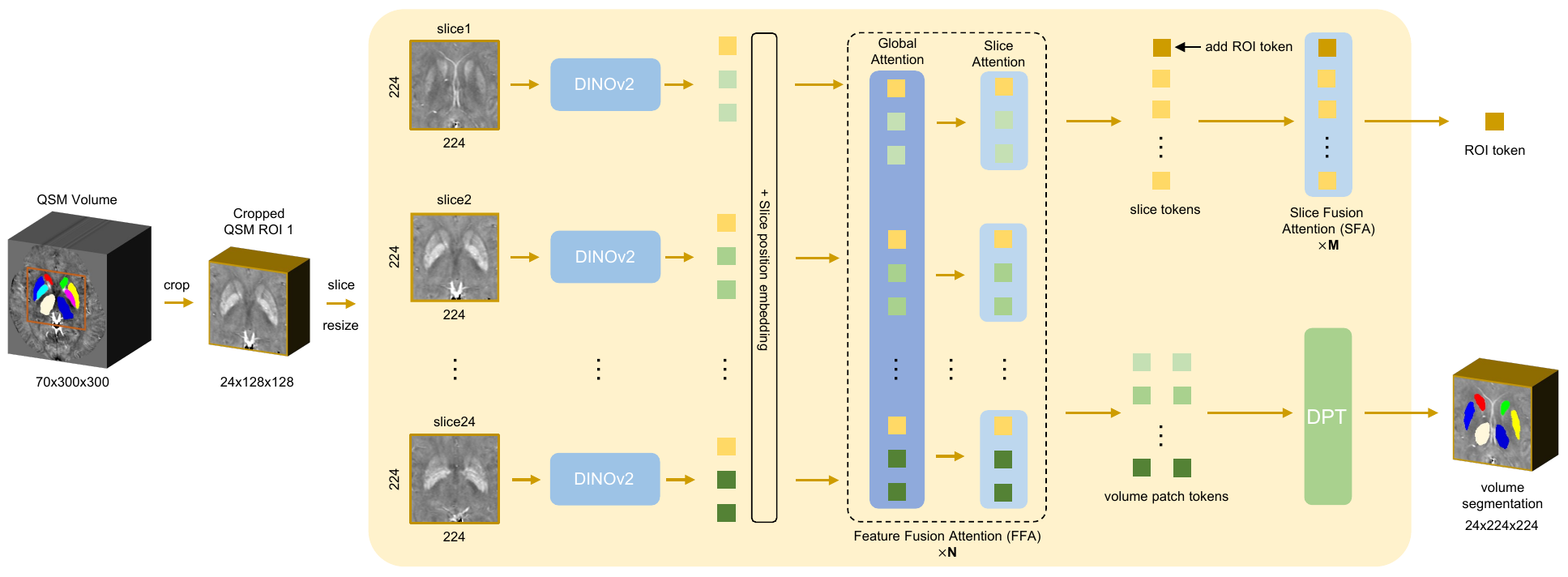}
    \caption{
    Overview of the \underline{R}OI volume feature \underline{E}xtraction and \underline{S}egmentation (RES). RES takes a cropped ROI as input and produces two outputs: 1) an ROI embedding that aggregates global volumetric information, and 2) a volume segmentation map that guides the feature extraction module to focus on the target region. To address the challenge of limited data, the last attention block of DINOv2 is reused to initialize all attention blocks in both the FFA and SFA modules.
    }
    \label{fig:res}
\end{figure}

As shown in Fig.\ref{fig:res}, our goal is to leverage DINOv2\cite{oquab2023dinov2} to extract a global ROI token from a cropped 3D ROI. 
An auxiliary segmentation branch is incorporated to guide the feature extraction branch to focus on the highlighted brain nuclei. 
Cropping the ROI also reduces the computational cost of self-attention compared to processing the entire 3D image.

Given a cropped 3D ROI volume $V \in \mathbbm{R}^{D \times H \times W}$ consisted of a set of 2D axial slices $(I_i)^D_{i=1}$,
RES serves as one mapping function as follows:
\begin{equation}
f((I_i)^D_{i=1}) = (p^{seg}, t^{ROI}),
\end{equation}
where $p^{seg}=(p_{i}^{seg})_{i=1}^{D}$ denotes the concatenated axial segmentation results of brain nuclei over all slices, and $t^{ROI} \in \mathbbm{R}^C$ is the global token representing the whole ROI volume\footnote{If no specification, all tokens have $C$ channels}.

Diving into the mapping function, we first use DINOv2 as a patchnizer to patchnify $i$-th 2D slice $I_i$ into local patch tokens $(t_j)_{j=1}^L$ as follows:
\begin{equation}
DINOv2(I_i) = ((t_j^i)_{j=1}^L, t^{slice}_i),
\end{equation}
where $t^{slice}_i$ is the slice token representing the slice local information, which is actually the inherent [CLS] token of DINOv2, and $L$ is the sequence length of the patch tokens\footnote{
Suppose patch size is $s$, slice of shape $H \times W$, then $L=\frac{H}{s} \times \frac{W}{s}$
}.

Then, inspired by VGGT~\cite{wang2025vggt}, we introduce a feature fusion self-attention (FFA) module, 
which alternates between global self-attention and local slice self-attention to model interactions among patch tokens and slice tokens at different levels. 
Specifically, the global attention aggregates 3D volumetric information within the ROI, 
while the local attention refines intra-slice details to prevent the global attention from confusing the spatial origins of patch tokens. 
We stack $N$ FFA blocks to further enhance its ability for hierarchical information aggregation.
Following FFA, all patch tokens are fed into a dense prediction transformer (DPT)~\cite{ranftl2021vision} to output the brain nuclei segmentation $p^{seg}$. 
Meanwhile, all slice tokens, along with an added ROI token, are passed into slice fusion self-attention (SFA) modules, 
which integrates global volumetric information to produce the final ROI token $t^{ROI}$. 
We repeat SFA for $M$ blocks to strengthen representation learning.

Different from VGGT, we add slice position embeddings $(pos^{slice}_i)_{i=1}^D \in \mathbbm{R}^{C}$ on all $D$ sets of patch tokens to let FFA and SFA aware of the consecutive depth relations.
It can be written as follows:
\begin{equation}
(\hat{t_j^i})_{j=1}^L = (t_j^i)_{j=1}^L + pos^{slice}_i, i=1,2,...,D.
\end{equation}

\subsubsection{Dense Prediction Transformers for Segmentation}
Traditionally, DPT~\cite{ranftl2021vision} takes patch tokens\footnote{$\hat{t_j^i}$ or $t_j^i$} from four different layers of DINOv2, 
converts them back into spatial feature maps, 
and progressively aggregates them to segment brain nuclei. 
In contrast, inspired by VGGT~\cite{wang2025vggt}, we primarily use patch tokens from the $N$ FFA layers, 
supplemented only by the final layer of DINOv2, to perform segmentation.

Specifically, a DPT layer first unpatchifies the patch tokens of a 2D slice $(t_j^i)_{j=1}^L \in \mathbbm{R}^{C}$ into a spatial feature map $(F^i_l)_{l=1}^N \in \mathbbm{R}^{C \times h \times w}$, where $l$ denotes the FFA layer index, $h=\frac{H}{s}$ and $w=\frac{W}{s}$, and $s$ is the DINOv2 patch size.
In this work, we set $N=3$ to reduce the risk of overfitting.
The model then progressively upsamples deep $F^i_{l+1}$ and fuses the upsampled deep features with shallow $F^i_l$ to generate a segmentation map $p_{i}^{seg}$ at one-quarter the original resolution for the $i$-th slice.
Finally, all slice-level maps are upsampled back to the original $H \times W$ resolution and concatenated to form the volumetric segmentation map $p^{seg}$.

\begin{figure}
    \centering
    \includegraphics[width=1.0\linewidth]{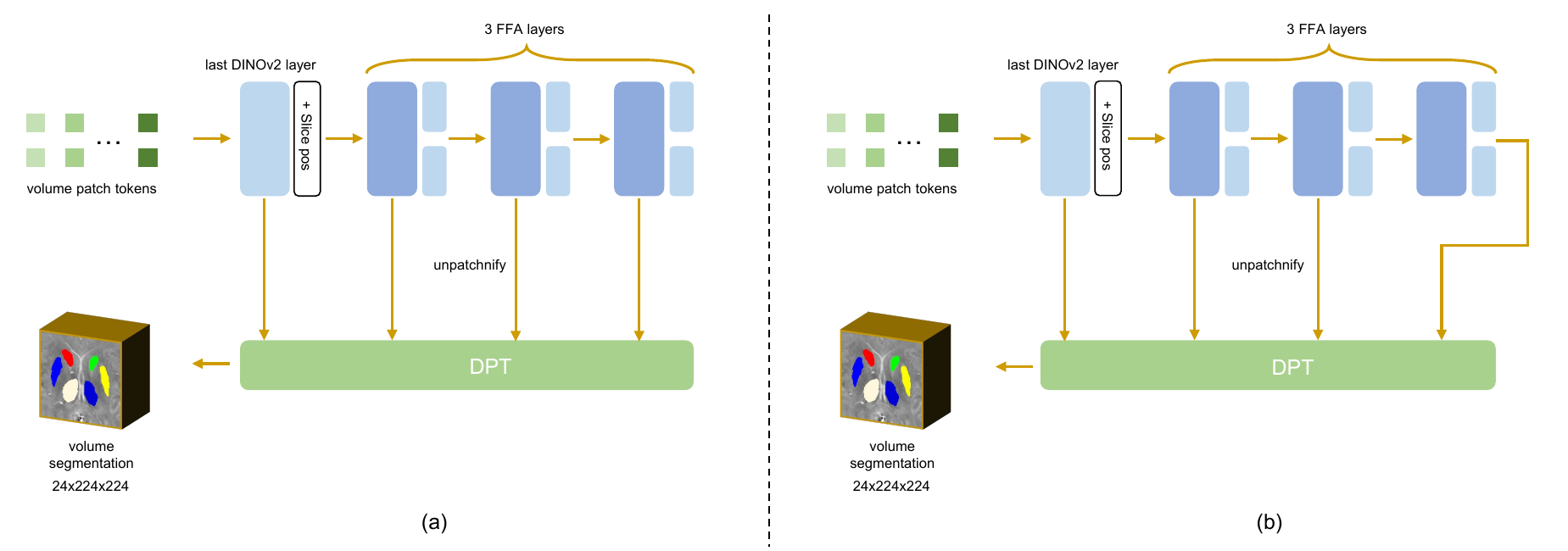}
    \caption{
    We explore two choices of patch tokens from the intermediate attention layers when setting $N=3$ in FFA:
    (a) Global: using volume tokens from the global attention of FFA layers, combined with the initial tokens from the last DINOv2 layer.
    (b) Local: identical to (a), except that the final patch tokens are taken from the local slice attention.
    Although approach (a) may seem more intuitive since we need globally attended patch tokens for volumetric segmentation, quantitative experiments demonstrate that approach (b) yields better performance (See Sec~\ref{tab:ab_dpt}).
    }
    \label{fig:dpt}
\end{figure}

As shown in Fig.~\ref{fig:dpt}, following~\cite{wang2025vggt, ranftl2021vision}, we send four intermediate tokens into DPT to perform volume segmentation.
With $N$=3, we design two choices for selecting patch tokens: 
\begin{itemize}
    \item \textit{Global}: We consider the tokens from the global attention layer of FFA since global information is important to 3D volume segmentation.
    \item \textit{Local}: All same as Global excepts the final patch tokens coming from the local slice attention of the final FFA layer.
\end{itemize}
We talk about which one is better in ablation studies.

\subsection{Multi-ROI Driven Classification Network}
\label{sec:mrn}
A straightforward approach to classify PD versus HC using paired NM and QSM images is to feed each whole image separately into RES, 
producing two global image tokens (replacing the ROI token with an image token) for the classification head. 
However, when the dataset is small, the resulting representations can be fragile, 
often causing the model to focus on irrelevant background regions rather than disease-relevant structures.

To this end, we propose MRN illustrated in Fig.~\ref{fig:mrn}.
MRN processes four ROIs from NM and QSM images using a shared RES module, 
encouraging the network to focus on disease-relevant foreground regions.
The brain nuclei corresponding to the four ROIs are listed as follows:
\begin{itemize}
    \item NM ROI $V^{NM} \in \mathbbm{R}^{12\times32\times64}$: substantia nigra.
    \item QSM ROI 1 $V^{QSM_1} \in \mathbbm{R}^{24\times128\times128}$:  caudate nucleus, putaman, globus pallidus, and thalamus.
    \item QSM ROI 2 $V^{QSM_2} \in \mathbbm{R}^{12\times64\times96}$: subthalamic nucleus, substantia nigra, and red nucleus.
    \item QSM ROI 3 $V^{QSM_3} \in \mathbbm{R}^{12\times48\times96}$: dentate nucleus.
\end{itemize}
The longest side of ROIs are resized to 224 using bi-linear interpolation.

With RES as the function $f$, we have the following equations to process the ROIs:
\begin{equation}
f(V^{NM}) = (p^{seg_{NM}}, t^{ROI_{NM}}),
\end{equation}
\begin{equation}
f(V^{QSM_k}) = (p^{seg_{QSM_k}}, t^{ROI_{QSM_k}}), k = 1,2,3.
\end{equation}

\subsubsection{Classification Head}\label{sec:cls_head} We concatenate all ROI tokens as the patient's representation, and feed it to a 2-layer MLPs with a dropout layer to predict the subject' class:
\begin{equation}
r=Concat(t^{ROI_{NM}}, t^{ROI_{QSM_1}}, t^{ROI_{QSM_2}}, t^{ROI_{QSM_3}}), r \in \mathbbm{R}^{4C},
\end{equation}
\begin{equation}
p^{cls} = MLP(r).
\end{equation}

\subsubsection{Multi-ROI Supervised Contrastive Pretraining} To further enhance the discriminability of representations between PD and HC, 
we employ mSupMoCo, which pulls together patient-level representations of the same class while pushing apart those of different classes~\cite{ding2025c3r, he2020momentum}. 
The motivation is two folds: 1) given the small dataset, irrelevant background in cropped ROIs may hinder learning. 
mSupMoCo encourages RES to extract generalizable representations with strong class differentiation by leveraging a large set of positive and negative samples; 
2) ROI cropping reduces the risk of overfitting to background but disrupts the spatial relationship between ROIs from the same patient. 
To address this, mSupMoCo operates on patient-level representations $r$, 
reconstructing the relations between ROIs via contrastive learning rather than treating each ROI independently.

Specifically, let $z \in \mathbbm{R}^{c}$ be the projection of the patient-level representation $r \in \mathbbm{R}^{4C}$ via a two-layer non-linear MLP, following C$^3$R~\cite{ding2025c3r} and MoCo~\cite{he2020momentum}. The contrastive loss is defined as:

\begin{equation}
\label{loss:cl}
L^{cl} = -\frac{1}{|P|} \sum_{p=1}^{|P|} log \frac{exp(z \cdot z^+_p / \tau)}{exp(z \cdot z^+_p / \tau) + \sum_{n=1}^{|N^-|}exp(z \cdot z_n^- / \tau)},
\end{equation}
where $z$ is the anchor sample, 
$\tau$ is a temperature controlling the strength of attraction and repulsion, 
$z^+_p$ is a positive sample from the same class as $z$, 
and $z^-_n$ is a negative sample from a different class. 
$|P|$ and $|N^-|$ denote the number of positive and negative samples in the minibatch. 
Here, $z$ is obtained from a student MRN with parameters $\theta_s^{MRN}$, 
while $z^+_p$ and $z^-_n$ are encoded by a teacher MRN with parameters $\theta_t^{MRN}$.

\paragraph{Category Memory Bank} Following~\cite{ding2025c3r, he2020momentum}, we build a memory bank $M$ of size $|M|$ for each class to cache stable patient representations during training.
The memory bank provides all samples cached with $P$ and $N^-$ to compensate the small training samples within the minibatch.
The bank is updated by a teacher MRN whose parameter updating rule is formated as follows:
\begin{equation}
\theta_t^{MRN} = m\theta_t^{MRN} + (1-m)\theta_s^{MRN},
\end{equation}
where $m$ is the hyperparameter controlling the updating velocity of the teacher MRN, and we empirically set it at 0.999.

\subsection{Training Objectives}
\label{sec:losses}
For segmenting brain nuclei of different ROIs in MRN, we have the following seg loss:
\begin{equation}
    \begin{aligned}
        L^{seg} & = WCE(p^{seg_{NM}}, y^{seg_{NM}}) + DICELoss(p^{seg_{NM}}, y^{seg_{NM}}) \\
                & + \sum_{k=1}^3 (WCE(p^{seg_{QSM_k}}, y^{seg_{QSM_k}}) + DICELoss(p^{seg_{QSM_k}}, y^{seg_{QSM_k}})),
    \end{aligned}
\end{equation}
where $WCE$ means weighted cross entropy loss.
We calculate the weights using 1 minus the class proportions within the image.
Note that, despite we show brain nuclei mask is distinct with left and right brains in figures, we combine left-right labels into one label to allow us to use aggressive left-right mirror augmentation.

For classification loss, we have cross-entropy loss as follows:
\begin{equation}
    L^{cls} = CE(p^{cls}, y^{cls}).
\end{equation}

The overall losses can be written as follows:
\begin{equation}
\label{loss:all}
    L^{all} = L^{cls} + \lambda^{seg}L^{seg} + \lambda^{cl}L^{cl},
\end{equation}
where $\lambda^{seg}$ and $\lambda^{cl}$ are trade-off hyperparameters weighting the importance of segmentation loss and contrastive loss, respectively.

\section{Experiment}
\subsection{Dataset and Implementation Details}

\subsubsection{MICCAI'2025 PDCADxFoundation Challenge Dataset} This dataset consists of 500 paired NM and QSM images~\cite{wang_2025_15094607}. 
Initially, 200 class-balanced samples (100 HC, 100 PD) are provided for training, 
along with diagnosis labels and segmentation masks; we refer to these as the training set. 
For validation, 100 balanced samples with labels and masks are released. 
The final testing set includes 200 class-balanced samples used for the evaluation of developed models.

For data splitting, we perform 5-fold cross-validation on the 200-sample training set to tune hyperparameters on the 100-sample validation set, using a fixed random seed of 1. 
Unless otherwise specified, all ablation studies are conducted on the first fold’s internal train/validation split and evaluated on the 100-sample validation set. 
After hyperparameter tuning, we merge the training and validation sets into 300 samples and apply 5-fold cross-validation for model ensembling, 
which is then submitted for the final test stage.

\subsubsection{Implementation Details} To learn disease representations that distinguish HC and PD while reducing overfitting, MRN is pretrained for 400 epochs using only the segmentation and contrastive losses. 
After pretraining, the model is trained for an additional 200 epochs using the full loss in Eq.~\ref{loss:all}, 
resulting in a total of 600 end-to-end training epochs.

We use the AdamW optimizer with betas of 0.9 and 0.999 and a batch size of 4. Different learning rates and weight decays are set for different modules:
\begin{itemize}
    \item DINOv2, FFA, SFA: learning rate 1e-5, weight decay 1e-1.
    \item DPT, classification head, contrastive head: learning rate 1e-4, weight decay 5e-2.
\end{itemize}
No adapters, LoRA, or visual prompts are used for DINOv2, as full-parameter tuning is preferred for the PDCAD task. 
Random augmentations include mirrors, rotations, intensity and contrast shifts, Gaussian noise, and histogram shifts.

For hyperparameters, we set number of layers of FFA $N$=3, and SFA layers $M$=2.
We set the temperature $\tau$ in contrastive loss at 0.1, which is selected from {0.07, 0.1}.
The size of memory bank $|M|$ is 1,024.
The dropout in the classification head is set at 0.5, and the drop path rate of the DINOv2 is set as 0.2.
We use ViT-B with four register tokens as the DINOv2 model.
And local manner as shown in Fig.~\ref{fig:dpt} for segmentation,
the project channels used in DPT~\cite{ranftl2021vision} are 96, 192, 384, and 768 and channels in the fusion layers of DPT is set as 256.

Since segmentation masks are unavailable during inference, we train NM and QSM segmentation models using nnUNetv2~\cite{isensee2021nnu} for 250 epochs on the 200 training samples (without mirror augmentations). 
During inference, auxiliary modules including DPT, contrastive head, momentum MRN, and the memory bank, are discarded.

MRN is developed on an NVIDIA H100 GPU (80 GB, decrease batch size from 4 to 1 will save the memory a lot) using PyTorch 2.2 and MONAI 1.3.0~\cite{cardoso2022monai}. 
Training 600 epochs takes approximately 6 hours. 
We perform 5-fold cross-validation for the merged 300 labeled samples in the final test phase, with predictions averaged across folds. 
Test-time augmentation (TTA) is only applied in final submissions, not in other experiments.

\subsubsection{Evaluation Metrics} We use area under the receiver operating curve (ROC\_AUC), accuracy, and F1-score to reflect the classification performance of our methods. 
F1-score is adopted mainly for evaluating the classification bias between HC and PD.

\subsection{A Strong CNN Baseline}
\label{sec:cnn_baseline}
To evaluate the effectiveness of our framework, we construct a strong baseline using a modified 3D SE-ResNet with 10 layers and SE blocks\cite{hu2018squeeze}, referred to as SERes10. 
This network is implemented using the built-in API in MONAI\cite{cardoso2022monai}. 
Unlike MRN, SERes10 does not employ multi-task learning. 
Instead, the one-hot segmentation mask of each ROI is concatenated to form a mask-enhanced ROI input.

For different input channels of the four ROIs defined in Sec.\ref{sec:mrn}, we build separate input stems and share the SERes10 backbone. 
The extracted 512-dimensional features from each ROI are combined to form the patient-level representation, 
which is then fed to the same classification head described in Sec.\ref{sec:cls_head}.

We also enhance this baseline with mSupMoCo pretraining, termed SERes10-mSupMoCo. 
Training this baseline is very fast (1 hour on an NVIDIA 2080 Ti), 
allowing extensive hyperparameter tuning as described in the implementation details.

\begin{table}
\centering
\caption{Comparison results with other top-5 solutions on the final test stage on 200 class-balanced samples.}\label{tab:cmpr_test}
\begin{tabular}{|l|l|l|l|}
\hline
Solution &  Accuracy & ROC\_AUC & F1-score \\
\hline
MRN (ours) &  86.0\% & N/A & N/A \\
2nd place &   80.5\% & N/A & N/A  \\
3rd place &   79.5\% & N/A & N/A \\
4th place &   76.5\% & N/A & N/A \\
5th place &   75.5\% & N/A & N/A \\
\hline
\end{tabular}
\end{table}

\subsection{Comparison results with other top solutions and Ablation study}
\subsubsection{Comparison Results}
As shown in Table~\ref{tab:cmpr_test}, we present the results on the final test set of 200 class-balanced samples. 
Our MRN achieves first-place accuracy, outperforming the second-best method (80.5\%) by approximately 5.5\%. 
These results demonstrate that the proposed MRN combined with RES effectively enables the analysis of 3D MR medical images.

\subsubsection{Ablation Study} We conduct comprehensive ablation studies to investigate the contribution of each component of MRN.
We first ablate components including FFA, SFA, DPT and the choice of its accepted features shown in Fig.~\ref{fig:dpt}, and mSupMoCo.
Then, we investigate the impact of number of FFA and SFA layers to the diagnostic performance.
Next, we wonder the contribution of the four ROIs and randomly mask them to simulate ROI missing during inference.
We also conduct ablation studies using the strong CNN baseline to clarify the hyperparameter impacts of mSupMoCo, including the size of the memory bank and the temperature. 
Note that we perform most ablation studies using the validation-100 dataset.

\begin{table}[]
\centering
\caption{Ablation studies of each component of MRN on the validation 100 samples. All results are obained with only single run using a fixed random seed 1 to speed up the experiments.}\label{tab:ab_contribution}
\begin{tabular}{|lllll|l|l|l|}
\hline
\multicolumn{5}{|l|}{Components}                                                                                         & \multicolumn{1}{c|}{\multirow{2}{*}{Accuracy}} & \multicolumn{1}{c|}{\multirow{2}{*}{ROC\_AUC}} & \multicolumn{1}{c|}{\multirow{2}{*}{F1-score}} \\ \cline{1-5}
\multicolumn{1}{|l|}{DINOv2} & \multicolumn{1}{l|}{FFA} & \multicolumn{1}{l|}{SFA} & \multicolumn{1}{l|}{DPT} & mSupMoCo & \multicolumn{1}{c|}{}                          & \multicolumn{1}{c|}{}                          & \multicolumn{1}{c|}{}                          \\ \hline
\multicolumn{1}{|l|}{\checkmark}      & \multicolumn{1}{l|}{}    & \multicolumn{1}{l|}{}    & \multicolumn{1}{l|}{}    &                   & 50.0\% & 50.0\%  & 33.3\%  \\ \hline
\multicolumn{1}{|l|}{\checkmark}      & \multicolumn{1}{l|}{\checkmark}   & \multicolumn{1}{l|}{}    & \multicolumn{1}{l|}{}    &          & 72.0\% & 79.3\%  & 72.0\%  \\ \hline
\multicolumn{1}{|l|}{\checkmark}      & \multicolumn{1}{l|}{}    & \multicolumn{1}{l|}{\checkmark}   & \multicolumn{1}{l|}{}    &          & 50.0\% & 50.0\%  & 33.3\%  \\ \hline
\multicolumn{1}{|l|}{\checkmark}      & \multicolumn{1}{l|}{}    & \multicolumn{1}{l|}{}    & \multicolumn{1}{l|}{\checkmark}   &          & 50.0\% & 50.0\%  & 33.3\%  \\ \hline
\multicolumn{1}{|l|}{\checkmark}      & \multicolumn{1}{l|}{\checkmark}   & \multicolumn{1}{l|}{\checkmark}   & \multicolumn{1}{l|}{}   &  & 55.0\% & 62.2\%  & 55.0\%   \\ \hline
\multicolumn{1}{|l|}{\checkmark}      & \multicolumn{1}{l|}{\checkmark}   & \multicolumn{1}{l|}{}    & \multicolumn{1}{l|}{\checkmark}   &  &  77.0\%   & 82.6\%  &  76.6\%        \\ \hline
\multicolumn{1}{|l|}{\checkmark}      & \multicolumn{1}{l|}{}    & \multicolumn{1}{l|}{\checkmark}   & \multicolumn{1}{l|}{\checkmark}   &  &  50.0\%   & 50.0\%  & 33.3\%         \\ \hline
\multicolumn{1}{|l|}{\checkmark}      & \multicolumn{1}{l|}{\checkmark}   & \multicolumn{1}{l|}{\checkmark}   & \multicolumn{1}{l|}{\checkmark}   &   & 75.0\% & 85.5\% &  74.9\%  \\ \hline
\multicolumn{1}{|l|}{\checkmark}      & \multicolumn{1}{l|}{\checkmark}   & \multicolumn{1}{l|}{\checkmark}   & \multicolumn{1}{l|}{\checkmark}   & \checkmark        &  83.0\% & 91.2\% & 83.0\%   \\ \hline
\multicolumn{1}{|l|}{\checkmark}      & \multicolumn{1}{l|}{\checkmark}   &  \multicolumn{1}{l|}{}  &  \multicolumn{1}{l|}{\checkmark}  & \checkmark   & 75.0\%  & 85.1\%  &  75.0\%  \\ \hline
\end{tabular}
\end{table}

\paragraph{Effects of each component of MRN}
We investigate the contribution of each component in MRN through ablation experiments, summarized in Table~\ref{tab:ab_contribution}. 
Below, we first clarify the ablation settings. 
For the “DINOv2” baseline, we directly average slice tokens for classification. 
For “DINOv2 + DPT,” we adopt the default DPT configuration, 
which uses patch tokens from layers 3, 6, 9, and 11 for segmentation.

Our results reveal several key insights. 
First, directly fine-tuning DINOv2 collapses to a PD-biased classifier, and even when augmented with DPT, 
it fails to achieve balanced classification. 
This indicates that trivial fine-tuning of 2D slices is not a feasible solution for the APD task. 
Among all modules, FFA serves as the key role: removing FFA leads MRN to again collapse into a biased classifier, 
highlighting the critical role of volume modeling for 3D image analysis. 
Adding DPT on top of FFA further improves accuracy (from 72.0\% to 77.0\%), 
showing that segmentation supervision steers the model to focus on disease-relevant regions.

It seems that SFA alone does not always improve accuracy. 
When combined with only DINOv2 and FFA, or with DINOv2, FFA, and DPT, accuracy drops from 72.0\% to 55.0\% and from 77.0\% to 75.0\%, respectively. 
However, its necessity becomes evident in the full MRN pipeline: removing SFA from the complete combination (DINOv2, FFA, SFA, DPT, mSupMoCo) causes the accuracy to decrease by 8.0\%, 
demonstrating that SFA is essential for effectively integrating information across slices once all other components are present.

\begin{table}[]
\centering
\caption{Ablation studies of the number of FFA and SFA layers on the validation 100 samples. All results are obained with only single run using a fixed random seed 1 to speed up the experiments.}\label{tab:ab_ffa_sfa}
\begin{tabular}{|ll|l|l|l|}
\hline
\multicolumn{2}{|l|}{Layers}     & \multicolumn{1}{c|}{\multirow{2}{*}{Accuracy}} & \multicolumn{1}{c|}{\multirow{2}{*}{ROC\_AUC}} & \multicolumn{1}{c|}{\multirow{2}{*}{F1-score}} \\ \cline{1-2}
\multicolumn{1}{|l|}{FFA} & SFA & \multicolumn{1}{c|}{}                          & \multicolumn{1}{c|}{}                          & \multicolumn{1}{c|}{}                          \\ \hline
\multicolumn{1}{|l|}{2}   & 1   &            83.0\%  &         89.8\%        &    83.0\%     \\ \hline
\multicolumn{1}{|l|}{2}   & 2   &            81.0\%  &         92.0\%        &    81.0\%     \\ \hline
\multicolumn{1}{|l|}{3}   & 1   &            81.0\%  &         90.6\%        &    81.0\%     \\ \hline
\multicolumn{1}{|l|}{3}   & 2   &            83.0\%  &         91.2\%        &    83.0\%     \\ \hline
\multicolumn{1}{|l|}{4}   & 1   &            83.0\%  &         87.0\%        &    82.8\%     \\ \hline
\multicolumn{1}{|l|}{4}   & 2   &            82.0\%  &         88.6\%        &    82.0\%     \\ \hline
\multicolumn{1}{|l|}{4}   & 3   &            84.0\%  &         88.6\%        &    83.9\%     \\ \hline
\end{tabular}
\end{table}

\paragraph{Effects of number of FFA and SFA layers}
As shown in Table~\ref{tab:ab_ffa_sfa}, we ablate the number of layers in FFA and SFA to evaluate their impact. 
Overall, the diagnostic performance remains relatively stable across different layer configurations. 
Although using 4 FFA layers and 3 SFA layers achieves slightly higher accuracy, 
we adopt 3 FFA layers and 2 SFA layers as a trade-off between computational cost and performance.

\begin{table}[]
\centering
\caption{Randomly masking ROIs of NM, QSM1, QSM2, and QSM3 to measure the contribution of each ROI to the diagnostic performance on the validation 100 samples. We use the MRN model trained using a fixed random seed 1. Sign \ding{53} denotes that we mask the given ROI in the column head.}\label{tab:ab_roi}
\begin{tabular}{|l|l|l|l|l|l|l|}
\hline
NM & QSM1 & QSM2 & QSM3 & \multicolumn{1}{c|}{Accuracy} & \multicolumn{1}{c|}{ROC\_AUC} & \multicolumn{1}{c|}{F1-score} \\ \hline
\multicolumn{4}{|l|}{baseline (No Masking)}                                               &    83.0\%    &   91.2\%    &  83.0\%   \\ \hline
\ding{53}  &      &      &                                      &    76.0\%    &   85.6\%    &  75.4\%   \\ \hline
   & \ding{53}    &      &                                      &    81.0\%    &   89.1\%    &  81.0\%   \\ \hline
   &      & \ding{53}    &                                      &    83.0\%    &   87.9\%    &  83.0\%     \\ \hline
   &      &      & \ding{53}                                    &    82.0\%    &   91.2\%    &  82.0\%     \\ \hline
\ding{53}  & \ding{53}    &      &                              &    72.0\%    &   82.4\%    &  71.1\%                             \\ \hline
\ding{53}  &      & \ding{53}    &                              &    65.0\%    &   73.7\%    &  63.4\%                          \\ \hline
\ding{53}  &      &      & \ding{53}                            &    68.0\%    &   84.9\%    &  66.7\%                         \\ \hline
  & \ding{53}    & \ding{53}    &                               &    74.0\%    &   81.0\%    &  74.0\%                             \\ \hline
& \ding{53}    &     &     \ding{53}                            &    82.0\%    &   90.9\%    &  82.0\%                        \\ \hline
&     &  \ding{53}   &     \ding{53}                            &    79.0\%    &   86.5\%    &  79.0\%                             \\ \hline
\ding{53}  & \ding{53}    & \ding{53}    &                      &    48.0\%    &   50.3\%    &  46.6\%                             \\ \hline
\ding{53}  & \ding{53}    &      & \ding{53}                    &    66.0\%    &   79.6\%    &  64.3\%                             \\ \hline
\ding{53}  &     &   \ding{53}   & \ding{53}                    &    61.0\%    &   76.6\%    &  57.9\%                             \\ \hline
   & \ding{53}    & \ding{53}    & \ding{53}                    &   72.0\%    &   82.6\%     &  72.0\%                             \\ \hline
\ding{53}  & \ding{53}    & \ding{53}    & \ding{53}            &    45.0\%    &    40.7\%  &   37.3\%                            \\ \hline
\end{tabular}
\end{table}

\paragraph{Contribution of different ROIs to the diagnostic performance} We further investigate the contribution of each ROI by randomly masking them during inference. 
This experiment also provides insight into the robustness of MRN when certain ROIs or MR modalities are missing. 
As shown in Table~\ref{tab:ab_roi}, the NM ROI emerges as the most critical for diagnosis. 
Accuracy drops by about 7\% when the NM ROI is masked while others remain available, and masking it together with any additional ROI leads to a substantial decline in performance. 
Interestingly, masking any single QSM ROI causes only a minor decrease in accuracy, 
suggesting that MRN is relatively robust to the absence of one QSM ROI.
This is importance to the clinical deployment because imperfect image pre-processing leads to missing QSM ROIs.
However, masking the QSM2 ROI, which corresponds to the midbrain region, 
along with other QSM ROIs results in a sharp decline in accuracy, 
highlighting the midbrain’s importance for distinguishing PD from HC.



\paragraph{Impact of hyperparameters in mSupMoCo} The number of negative samples is a key factor affecting contrastive learning performance~\cite{ding2025c3r, he2020momentum}. 
As shown in Table~\ref{tab:ab_mocoK}, we ablate the memory bank size $|M|$ using the CNN baseline to evaluate its effect. 
Classification accuracy stably improves as the memory bank size increases.
We also ablate the temperature $\tau$ of mSupMoCo in Eq.\ref{loss:cl} (Table\ref{tab:ab_mocoT}). 
Although $\tau=0.07$ achieves a slightly higher accuracy of 82.0\%, 
its training fails if we change the data splits and random seeds.
In contrast, $\tau=0.1$ provides stable training regardless of seed or data split and consistently improves classification performance.

\begin{table}
\centering
\caption{Impact of the size of memory bank in mSupMoCo using SERes10-2MLP on the validation 100 samples. All results are obained with only single run using a fixed random seed 1 to speed up the experiments. We use $tau$=0.07 in these experiments.}\label{tab:ab_mocoK}
\begin{tabular}{|l|l|l|l|}
\hline
Method &  Accuracy & ROC\_AUC & F1-score \\
\hline
SERes10 (3D CNN baseline) &  79.0\% & 88.1\% & 78.9\% \\
\hline
$|M|$=128 & 76.0\% & 85.4\% & 76.0\% \\
$|M|$=256 & 73.0\% & 84.5\% & 72.9\% \\
$|M|$=512 & 81.0\% & 89.9\% & 81.0\% \\
$|M|$=1024 & 82.0\% & 90.0\% & 82.0\% \\
\hline
\end{tabular}
\end{table}

\begin{table}
\centering
\caption{Importance of the temperature $tau$ in mSupMoCo using SERes10-2MLP on the class-balanced validation 100 samples. All results are obained with only single run using a fixed random seed 1 to speed up the experiments. We use $|M|$=1024 in these experiments. We select $tau$=0.1 finally as its training is more stable.}\label{tab:ab_mocoT}
\begin{tabular}{|l|l|l|l|}
\hline
Method &  Accuracy & ROC\_AUC & F1-score \\
\hline
SERes10 (3D CNN baseline) &  79.0\% & 88.1\% & 78.9\% \\
\hline
$tau$=0.01 & 80.0\% & 88.3\% & 80.0\% \\
$tau$=0.07 & 82.0\% & 90.0\% & 82.0\% \\
$tau$=0.1 & 81.0\% & 87.7\% & 81.0\% \\
\hline
\end{tabular}
\end{table}

\begin{table}
\centering
\caption{Ablations of different DINOv2 backbones. All results are obained with only single run using a fixed random seed 1 to speed up the experiments.}\label{tab:ab_vit}
\begin{tabular}{|l|l|l|l|}
\hline
Method &  Accuracy & ROC\_AUC & F1-score \\
\hline
SERes10-mSupMoCo &  81.0\% & 87.7\% & 81.0\% \\
\hline
MRN-ViT-Small &  80.0\% & N/A & 80.0\% \\
MRN-ViT-Base (ours) & 83.0\% & 91.2\% & 83.0\% \\
\hline
\end{tabular}
\end{table}

\paragraph{Ablations of different DINOv2 backbones} The backbone size of DINOv2 in RES affects classification performance. 
As shown in Table~\ref{tab:ab_vit}, upgrading from a small ViT to the base variant improves accuracy and F1-score by approximately 3.0\%, 
indicating that increasing model capacity enhances performance.

\begin{table}
\centering
\caption{Ablations of different choices of input features for DPT. All results are obained with only single run using a fixed random seed 1 to speed up the experiments.}\label{tab:ab_dpt}
\begin{tabular}{|l|l|l|l|}
\hline
Method &  Accuracy & ROC\_AUC & F1-score \\
\hline
SERes10-mSupMoCo &  81.0\% & 87.7\% & 81.0\% \\
\hline
MRN-DPT-Global &   82.0\% & N/A &   81.8\% \\
MRN-DPT-Local (ours) & 83.0\% & 91.2\% & 83.0\% \\
\hline
\end{tabular}
\end{table}

\paragraph{Ablations of using Global or Local in DPT} The DPT guides MRN to focus on specific brain nuclei for classification. 
Figure~\ref{fig:dpt} illustrates the two options to select the four features fed into DPT. 
Intuitively, the Global approach seems preferable, as the model should consider the entire 3D volume for diagnosis. 
However, as shown in Table~\ref{tab:ab_dpt}, switching from Global to Local improves classification accuracy by 1.0\%. 
This suggests that there is still potential to better integrate slice-level information into full 3D representations by designing appropriate proxy tasks.
Another reason may be that the classification accepts tokens from the local slice attention of the last FFA layer.

\section{Discussion and Conclusion}
\label{sec:discussion}
APD has attracted increasing attention due to the high prevalence of PD and its impact on patient prognosis. 
Current methods leverage imaging biomarkers from QSM and NM-MR images, highlighting their diagnostic potential~\cite{he2021imaging}. 
In this work, we harness 2D vision foundation models (VFMs) to perform APD using paired 3D QSM and NM images. 
To adapt 2D VFMs to 3D volumes, we use DINOv2 to extract features from multiple slices and introduce FFA and SFA modules to fuse slice-level features into a 3D representation. 
We further propose the MRN framework, which processes four key ROIs from QSM and NM images in parallel, compresses each into ROI tokens, and combines them into a patient-level representation for classification. 
On the MICCAI 2025 PDCADxFoundation challenge, MRN achieves first-place accuracy of 86.0\%, outperforming the second-best solution by approximately 5.5\%. 
Experimental results demonstrate the potential of applying 2D VFMs for analyzing 3D MR images.







Harnessing VFMs for APD using multi-parametric MRIs is a promising approach in clinical scenarios. 
Early methods relied on T2 images from public datasets such as PPMI~\cite{marek2018parkinson} to develop CNN-based APD models, 
achieving high accuracies above 88.0\%\cite{sivaranjini2020deep, kaur2021diagnosis}. 
However, these datasets are small (fewer than 200 samples), raising concerns about generalization to unseen images. 
Later methods focused on specific brain nuclei in QSM images using either radiomics\cite{li20193d, zhang2022histogram, kang2022combining} or deep learning~\cite{xiao2019quantitative, wang2023automatic}. 
Despite high reported diagnostic performance, they face similar overfitting and generalization issues.
In clinical practice, data resources are inherently limited. 
VFMs, pretrained on millions or billions of images, reduce data requirements when transferred to new tasks, 
often matching or surpassing the performance of models trained on much larger labeled datasets~\cite{wang2025vggt, rui2025multi}. 
Our proposed MRN provides a promising framework to leverage 2D VFMs for APD using only hundreds of NM and QSM images.

\begin{table}
\centering
\caption{Results of directly finetuning 2D or 3D VFMs compared to the CNN baseline on the validation 100 samples. All results are obained with only single run using a fixed random seed 1 to speed up the experiments.}\label{tab:finetune}
\begin{tabular}{|l|l|l|l|l|}
\hline
Method &  Data dimension &  Accuracy & ROC\_AUC & F1-score \\
\hline
SERes10 (Train from scratch) & 3D &  79.0\% & 88.1\% & 78.9\% \\
\hline
M3D-CLIP-ViT-Base & 3D &   70.0\% & 75.7\% &   70.0\% \\
DINOv2-ViT-Base & 2D  &   50.0\% & 50.0\%  & 33.3\% \\
MRN-ViT-Small   & 2D    & 80.0\% & N/A & 80.0\% \\
MRN-ViT-Base (ours) & 2D & 83.0\% & 91.2\% & 83.0\% \\
\hline
\end{tabular}
\end{table}

We observed that directly fine-tuning pretrained VFMs on the challenge dataset is not straightforward. 
As shown in Table~\ref{tab:finetune}, we trained M3D-CLIP\cite{bai2024m3d}, pretrained on 3D CT images and associated medical text reports in a CLIP-like fashion\cite{radford2021learning}, 
and a ViT-Base initialized with DINOv2~\cite{oquab2023dinov2}. 
Both models, when directly fine-tuned, failed to surpass the CNN baseline SERes10. 
This suggests that data misalignment between pretraining and fine-tuning contributes to the suboptimal performance.
In contrast, our MRN, leveraging the small DINOv2 ViT, consistently outperforms both the baseline CNN and the directly fine-tuned models. 
Moreover, performance improves further when scaling the ViT from small to base, highlighting the benefit of larger model capacity without the concern of overfitting.

Despite its promising performance, our method has several limitations that warrant future investigation.
First, MRN’s integration of DINOv2 with DPT, FFA, and SFA incurs substantial memory and computational overhead. 
While these modules improve classification, future work could explore ablating them and directly processing volume patch tokens using the native self-attention of DINOv2. 
Techniques such as SegFormer~\cite{xie2021segformer} could also be employed to downsample tokens along the depth axis to reduce computation.
Second, we have not tested whether scaling from a base ViT to larger models further improves accuracy due to the constraint of the heavy overhead of computation resource.
Future studies should first reduce memory and computation costs before exploring larger backbones.
Third, the dataset in this study is relatively small and collected from a single center. 
Expanding the dataset across multiple centers could better demonstrate MRN’s generalizability.
Finally, our method relies on multiple ROIs and imaging modalities. 
Future work should assess the individual contribution of each ROI and modality to classification performance.

In conclusion, we propose MRN, a framework that leverages 2D VFMs to automatically diagnose Parkinson’s disease using paired 3D QSM and NM images.
MRN achieves 86.0\% classification accuracy, surpassing other top solutions in the MICCAI 2025 PDCADxFoundation Challenge and outperforming a strong 3D CNN baseline trained from scratch. 
We analyze key components of MRN, including contrastive learning hyperparameters, small versus base ViT backbones, and the selection of FFA features for DPT. 
Additionally, we evaluate direct fine-tuning of 2D/3D VFMs and discuss their limitations, 
as well as the limitations and potential improvements of MRN itself.

\begin{credits}
\subsubsection{\ackname} A bold run-in heading in small font size at the end of the paper is
used for general acknowledgments, for example: This study was funded
by X (grant number Y).

\subsubsection{\discintname}
The authors declare that they have no known competing financial interests or personal relationships that could have appeared to influence the work reported in this paper.
\end{credits}
%
%
%
%






\printbibliography

\end{document}